# The University of California San Francisco Preoperative Diffuse Glioma MRI (UCSF-PDGM) Dataset

Evan Calabrese[1], Javier Villanueva-Meyer[1], Jeffrey Rudie[1], Andreas Rauschecker[1], Ujjwal Baid[2], Spyridon Bakas[2], Soonmee Cha[1], John Mongan[1], Christopher Hess[1]

[1] Center for Intelligent Imaging (Ci$^2$), University of California San Francisco, Department of Radiology & Biomedical Imaging
[2] University of Pennsylvania, Center for Biomedical Image Computing and Analytics (CBICA)

Corresponding Author
Evan Calabrese, MD, PhD
evan.calabrese@ucsf.edu

# 1 Introduction

MRI-based artificial intelligence (AI) research on patients with brain gliomas has been rapidly increasing in popularity in recent years in part due to a growing number of publicly available MRI datasets. Notable examples include The Cancer Imaging Archive's glioblastoma dataset (TCGA-GBM) consisting of 262 subjects and the International Brain Tumor Segmentation (BraTS) challenge dataset consisting of 542 subjects (including 243 preoperative cases from TCGA-GBM) (1–4). The public availability of these glioma MRI datasets has fostered the growth of numerous emerging AI techniques including automated tumor segmentation, radiogenomics, and survival prediction. Despite these advances, existing publicly available glioma MRI datasets have been largely limited to only 4 MRI contrasts (T2, T2/FLAIR, and T1 pre- and post-contrast) and imaging protocols vary significantly in terms of field strength and acquisition parameters.

Here we present the University of California San Francisco Preoperative Diffuse Glioma MRI (UCSF-PDGM) dataset. The UCSF-PDGM dataset includes 500 subjects with histopathologically-proven diffuse gliomas who were imaged with a standardized 3 Tesla preoperative brain tumor MRI protocol featuring predominantly 3D imaging including diffusion and perfusion imaging. The dataset also includes isocitrate dehydrogenase (IDH) mutation status for all cases and O[6]-methylguanine-DNA methyltransferase (MGMT) promotor methylation status for World Health Organization (WHO) grade 3 and 4 gliomas. Finally, we have also included treatment details including extent of resection and overall survival. The UCSF-PDGM has been made publicly available in the hopes that researchers around the world will use these data to continue to push the boundaries of AI applications for diffuse gliomas.

# 2 Methods

## 2.1 Patient Population
Data collection was performed in accordance with relevant guidelines and regulations and was approved by the University of California San Francisco institutional review board with a waiver for consent. The dataset population consisted of 500 adult patients with histopathologically confirmed grade 2-4 diffuse gliomas who underwent preoperative MRI, initial tumor resection, and tumor genetic testing at a single medical center between 2015 and 2021. Patients with any prior history of brain tumor treatment were excluded; however, prior tumor biopsy was allowed.

## 2.2 Surgical Treatment and Survival Data
Extent of resection and overall survival were determined by review of the electronic medical record. When available, extent of resection was based on the operative report and/or immediate postoperative MRI report. Overall survival was recorded in days from initial diagnosis to the date of death or last clinical follow up.

## 2.3 Genetic Biomarker Testing
All subjects' tumors were tested for *IDH* mutations by genetic sequencing of tissue acquired at biopsy or resection. All grade 3 and 4 tumors were tested for MGMT methylation status using a methylation sensitive quantitative PCR assay.

## 2.4 Image Acquisition
All preoperative MRI was performed on a 3.0 tesla scanner (Discovery 750, GE Healthcare, Waukesha, Wisconsin, USA) and a dedicated 8-channel head coil (Invivo, Gainesville, Florida, USA). The imaging protocol included 3D T2-weighted, T2/FLAIR-weighted, susceptibility-weighted (SWI), diffusion-weighted (DWI), pre- and post-contrast T1-weighted images, 3D arterial spin labeling (ASL) perfusion images, and 2D 55-direction high angular resolution diffusion imaging (HARDI). Acquisition parameters for each sequence are more completely described in prior publications (5). Over the study period, two gadolinium-based contrast agents were used: gadobutrol (Gadovist, Bayer, LOC) at a dose of 0.1 mL/kg and gadoterate (Dotarem, Guerbet, Aulnay-sous-Bois, France) at a dose of 0.2 mL/kg.

## 2.5   Image Pre-Processing

HARDI data were eddy current corrected and processed using the Eddy and DTIFIT modules from FSL 6.0.2 yielding isotropic diffusion weighted images (DWI) and quantitative maps: mean diffusivity (MD), axial diffusivity (AD), radial diffusivity (RD), and fractional anisotropy (FA) (6,7). Eddy correction was performed with outlier replacement. DTIFIT was performed with simple least squares regression. Each image contrast was registered and resampled to the 1 mm isotropic resolution 3D space defined by the T2/FLAIR image using automated non-linear registration (Advanced Normalization Tools) with previously published parameters (5,8). Resampled co-registered data were then skull stripped using a publicly available method (5,8): https://www.github.com/ecalabr/brain_mask/.

## 2.6   Tumor Segmentation

Multicompartment tumor segmentation of study data was undertaken as part of the 2021 BraTS challenge as previously described (1). Briefly, image data first underwent automated segmentation using an ensemble model consisting of prior BraTS challenge algorithms. Images were then manually corrected by a group of annotators with varying experience and approved by one of two neuroradiologists with > 15 years attending experience. Segmentation included three major tumor compartments: enhancing tumor, central non-enhancing/necrotic tumor, and surrounding FLAIR abnormality (consisting of non-enhancing tumor and associated edema).

## 3   Results

### 3.1   Study participant demographic data

Basic demographic data for all study participants are presented in Table 1. The 500 cases included in the UCSF-PDGM includes 55 (11%) grade 2, 42 (9%) grade 3, and 403 (80%) grade 4 tumors. There was a male predominance for all tumor grades (56%, 60%, and 60%, respectively for grades 2-4). IDH mutations were identified in a majority of grade 2 (83%) and grade 3 (67%) tumors and a small minority of grade 4 tumors (8%). MGMT promoter hypermethylation was detected in 63% of grade 4 gliomas. 1p/19q codeletion was detected in 20% of grade 2 tumors and a small minority of grade 3 (5%) and 4 (<1%) tumors.

*Table 1 - Study population demographics and tumor genetic characteristics by World Health Organization tumor grade (2-4). Race and ethnicity data was not available for the study population.*

| Parameter | All Grades | Grade 2 | Grade 3 | Grade 4 |
|---|---|---|---|---|
| **Total Number** | 500 | 55 | 43 | 400 |
| **Male** | 298/500 | 31/55 | 26/43 | 241/400 |
| **Female** | 201 (40%) | 24/55 | 17/43 | 159/400 |
| **Age** | 57 ± 15 | 41 ± 14 | 46 ± 14 | 60 ± 14 |
| **IDH Mutant** | 106/500 | 46/55 | 29/43 | 30/400 |
| **MGMT Methylated** | 262/411 | 5/7 | 15/22 | 242/381 |
| **1p/19q Co-deletion** | 16/408 | 11/55 | 2/43 | 2/309 |

### 3.2   Surgical Treatment and Survival Data

Surgical treatment and survival data are included for the entire study cohort. Figure 1 shows overall survival for glioblastoma cases in the cohort stratified by extent of resection.

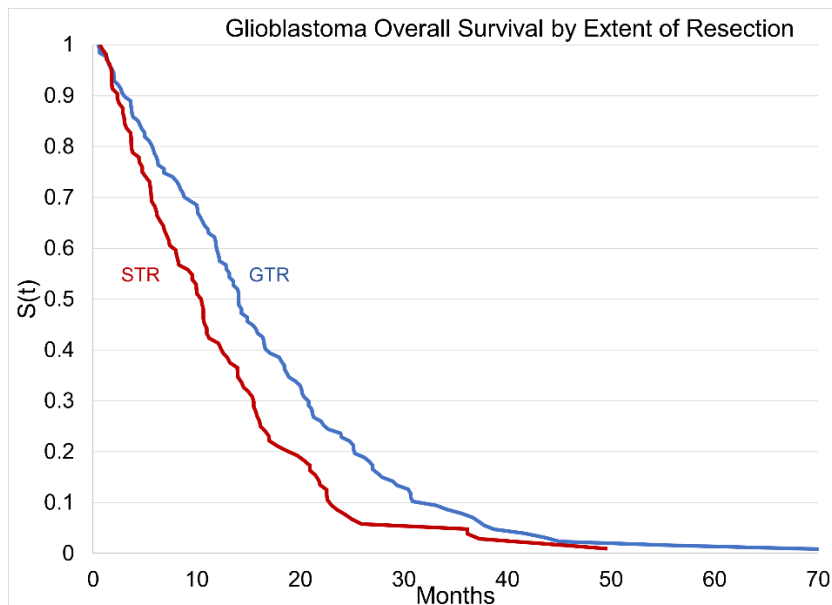

*Figure 1* – Overall survival of UCSF-PDGM patients with glioblastoma as a function of time (S(t)) stratified by extent of resection: gross total resection (GTR) versus subtotal resection (STR).

### 3.3 MR image data

A representative set of images from a single UCSF-PDGM subject is presented in Figure 2. Each subject includes skull-stripped co-registered 3D images in 11 different MRI contrasts as well as multicompartment tumor segmentations.

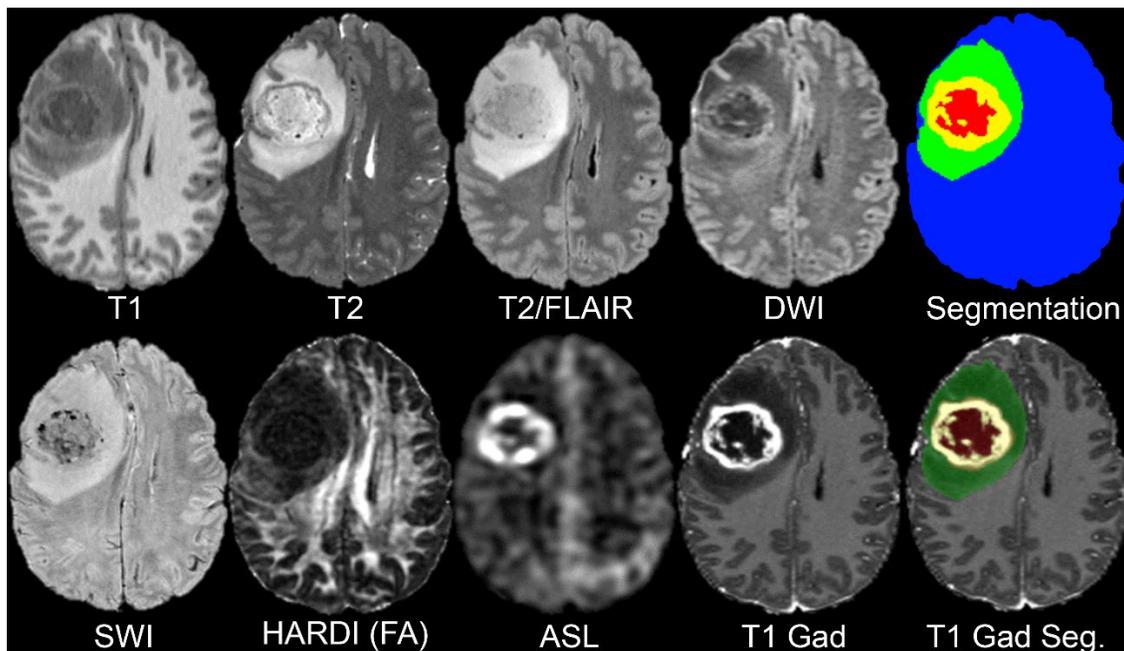

*Figure 2 - Representative multimodal MR images from a single case in the UCSF-PDGM dataset. T1: T1-weighted pre-contrast, T2: T2-weighted, T2/FLAIR: T2-weighted FLAIR, DWI: isotropic (trace) diffusion weighted image, SWI: susceptibility-weighted image, HARDI (FA): fractional anisotropy derived from HARDI data, ASL: arterial spin labeling perfusion, T1 Gad: T1-weighted post-contrast, Segmentation: multicompartment tumor segmentation (blue = brain,*

*green = FLAIR abnormality, yellow = enhancing tumor, red = necrotic core), T1 Gad Seg.: tumor segmentation semitransparent overlay on T1-weighted post-contrast image.*

### 3.4 Comparison to related datasets

Comparison of the UCSF-PDGM with similar existing resources is presented in Table 2. Comparison datasets include BraTS and TCGA as well as CPTAC-GBM, QIN-GBM, ACRIN-FMISO-Brain, and Ivy GAP (9–14). Notable differences include a higher number of cases, consistent 3 Tesla MR imaging protocol, and increased number of sequences.

*Table 2 - Comparison of selected publicly available preoperative diffuse glioma MRI datasets. * Training and validation cases only. Includes 243 cases from TCGA. † Excludes DWI and HARDI sequences, which are 2D. ‡ At least 1 genetic biomarker is provided. Genetic data not available for all patients.*

| Dataset | Cases | Tumor Grade | MRI Contrasts | Field Strength | Acquisition dimension | Segmentation Data | Genetic Data |
|---|---|---|---|---|---|---|---|
| **UCSF-PDGM** | 500 | 2-4 | T1, T1c, T2, FLAIR, DWI, SWI, HARDI, ASL | 3T | 3D † | Included | Included ‡ |
| **BraTS 2020** | 494* | 2-4 | T1, T1c, T2, FLAIR | 1.5T, 3T | 2D and 3D | Included | Not Included |
| **TCGA-GBM** | 262 | 4 | T1, T1c, T2, FLAIR | 1.5T, 3T | 2D and 3D | Not included | Included ‡ |
| **TCGA-LGG** | 199 | 2-3 | T1, T1c, T2, FLAIR | 1.5T, 3T | 2D and 3D | Not included | Included ‡ |
| **CPTAC-GBM** | 66 | 4 | Variable | 1.5T, 3T | 2D and 3D | Not included | Included ‡ |
| **QIN GBM** | 54 | 4 | T1, T2, FLAIR, MEMPRAGE, DWI, DCE | 3T | 2D and 3D | Not Included | Not Included |
| **ACRIN-FMISO-Brain** | 45 | 4 | T1, T1c, T2, FLAIR, DWI, DCE, DSC | 1.5T, 3T | 2D and 3D | Not included | Included ‡ |
| **Ivy GAP** | 39 | 4 | Variable | 1.5T, 3T | 2D and 3D | Not included | Included ‡ |

### 3.5 Data availability

As of July 2, 2021, a portion of the UCSF-PDGM dataset is available via the 2021 RSNA/ASNR/MICCAI BraTS challenge (https://www.med.upenn.edu/cbica/brats2021/). The entire UCSF-PDGM dataset will be publicly available via The Cancer Imaging Archive (TCIA) website (https://www.cancerimagingarchive.net/).

## 4 Discussion

The UCSF-PDGM adds to on an existing body of publicly available diffuse glioma MRI datasets that can be used in AI research. As MRI-based AI research applications continue to grow, new data are needed to foster development of new techniques and increase the generalizability of existing algorithms. The UCSF-PDGM not only significantly increases the total number of publicly available diffuse glioma MRI cases, but also provides a unique contribution in terms of MRI technique. The inclusion of 3D sequences and advanced MRI techniques like ASL and HARDI provides a new opportunity for researchers to explore the potential utility of cutting-edge imaging for AI applications.

The UCSF-PDGM dataset, particularly when combined with existing publicly available datasets, has the potential to fuel the next phase of radiologic AI research on diffuse gliomas. However, the UCSF-PDGM dataset's potential will only be realized if the radiology AI research community takes advantage of this new data resource. We hope that this dataset sparks inspiration in the next generation of AI researchers, and we look forward to the new techniques and discoveries that the UCSF-PDGM will generate.

## 5 Summary Statement

The University of California San Francisco Preoperative Diffuse Glioma MRI (UCSF-PDGM) dataset is a new publicly available brain MRI dataset consisting of 500 patients with grade 2-4 diffuse gliomas. The UCSF-PDGM data includes a standardized 3 Tesla, 3-dimensional, preoperative MR imaging protocol, diffusion and perfusion MRI, multicompartment tumor segmentations, tumor genetic data, and treatment/survival data.

# 6 Acknowledgment

Work reported in this publication was partly supported by the National Institutes of Health (NIH) under awards number NCI:U01CA242871 and T32EB001631 as well as by the Radiological Society of North America Research & Education (RSNA R&E) Foundation under award number RR2011. The content of this publication is solely the responsibility of the authors and does not necessarily represent the official views of the NIH or RSNA R&E Foundation.